\documentclass[11pt,a4paper]{article}
\usepackage[hyperref]{emnlp2020}
\usepackage{times}
\usepackage{latexsym}
\usepackage{graphicx}
\usepackage{float}
\usepackage{subcaption,booktabs}
\usepackage{microtype}
\usepackage{threeparttable}

\aclfinalcopy 

\title{BANANA at WNUT-2020 Task 2: Identifying COVID-19 Information on Twitter by Combining Deep Learning and Transfer Learning Models}

\author{  Tin Van Huynh \\
  University of Information Technology \\
  VNU-HCM, Vietnam \\
  {\tt 16521827@gm.uit.edu.vn} \\ \And
  Luan Thanh Nguyen \\
  University of Information Technology \\
  VNU-HCM, Vietnam \\
  {\tt 17520721@gm.uit.edu.vn}
  \\\AND 
  Son T. Luu \\
  University of Information Technology \\
  VNU-HCM, Vietnam \\
  {\tt sonlt@uit.edu.vn}\\}

\date{}

\begin{document}
\maketitle
\begin{abstract}
The outbreak COVID-19 virus caused a significant impact on the health of people all over the world. Therefore, it is essential to have a piece of constant and accurate information about the disease with everyone. This paper describes our prediction system for WNUT-2020 Task 2: Identification of Informative COVID-19 English Tweets. The dataset for this task contains size 10,000 tweets in English labeled by humans. The ensemble model from our three transformer and deep learning models is used for the final prediction. The experimental result indicates that we have achieved F1 for the INFORMATIVE label on our systems at 88.81\% on the test set.
\end{abstract}

\section{Introduction}
\label{introduction}
The rapid spread of the coronavirus (COVID-19) has caused a global health crisis. This virus is hazardous to people's health and causes a big panic all over the world. Statistics show that each day there are 4 million tweets related to COVID-19 on Twitter \cite{781w-ef42-20}. Therefore, it is essential to keep track of the information associated with this disease. Along with the development of many social networking platforms such as Twitter and Facebook. This is the primary way that helps people capture information about COVID-19 regularly. However, there is much content appearing daily on these social media platforms. Most of them do not have information about the status of COVID-19, such as the number of suspected cases or cases near the user's area.

In this article, we present our approach at WNUT-2020 Task 2 \cite{covid19tweet} to identify Tweets containing information about COVID-19 on the social networking platform Twitter or not. A Tweet is believed to have information if it includes information such as recovered, suspected, confirmed, and death cases and location or travel history of the patients. Specifically, we described the problem as follows.

\begin{itemize}
    \item \textbf{Input}: Given English Tweets on the social networking platform.
    \item \textbf{Output}: One of two labels (INFORMATIVE and UNINFORMATIVE) predicted by our system.
\end{itemize}
Several examples are shown in Table \ref{tab:example}

\begin{table}[H]
\begin{tabular}{p{6cm}c}
\hline
\multicolumn{1}{c}{\textbf{Tweet}} &
  \multicolumn{1}{l}{\textbf{Label}} \\ \hline
A New Rochelle rabbi and a White Plains doctor are among the 18 confirmed coronavirus cases  in Westchester. HTTPURL &
  0 \\ \hline
Day 5: On a family bike ride to pick up dinner at @USER Broadway, we encountered our pre-COVID-19 Land Park happy hour crew keeping up the tradition at an appropriate \#SocialDistance.HTTPURL &
  1 \\ \hline
\end{tabular}
\caption{Several examples in the WNUT-2020 Task 2 dataset. 0 and 1 stand for INFORMATIVE and  UNINFORMATIVE, respectively.}
\label{tab:example}
\end{table}

In this paper, we have two main contributions as follows.
\begin{itemize}
    \item Firstly, we implemented four different models based on neural networks and transformers such as Bi-GRU-CNN, BERT, RoBERTa, XLNet to solve the WNUT-2020 Task 2: Identification of informative COVID-19 English Tweets. \newline
    \item Secondly, we propose a simple ensemble model by combining multiple deep learning and transformer models. This model gives the highest performance compared with the single models with F1 on the test set is 88.81\% and on the development set is 90.65\%.
\end{itemize}

\section{Related work}
\label{relatedworks}
During the happening of the COVID-19 pandemic, the information about the number of infected cases, the number of patients is vital for governments. \citet{en-dong-lancet} constructed a real-time database for tracking the COVID-19 around the world. This dataset is collected by experts from the World Health Organization (WHO), US CDC, and other medical agencies worldwide and is operated by John Hopkins University. Also, there are many other COVID-19 datasets such as multilingual data collected on Twitter from January 2020 \cite{chen2020tracking} or Real World Worry Dataset (RWWD) \cite{kleinberg2020measuring}.

Besides, on social media, the spreading of COVID-19 information is extremely fast and enormous and sometimes leads to misinformation. \citet{shahi2020exploratory} conducted a pilot study about detecting misinformation about COVID-19 on Twitter by analyzing tweets using standard social media analytics techniques. From the researching results, the authors want to help authorities and social media users counter misinformation. Moreover, the rumors and conspiracy theories within the emergence times of COVID-19 spreading had made communities feel fearmongering and panicky, which lead to racism about COVID-19 patients and citizens from infected countries, and mass purchase of face masks as well as the shortage of necessaries, according to \cite{10.1093/jtm/taaa031}. Thus it is necessary to identify the right information from the social media text.
\section{Dataset}
\label{dataset}

The dataset provided by \citet{covid19tweet} contains 10,000 English Tweets about COVID-19, which is used to automatically identify whether a tweet contains useful information about the COVID-19 (informative) or not (uninformative). There are 4,719 INFORMATIVE tweets and 5,281 UNINFORMATIVE tweets in the dataset, and three different annotators annotate each tweet. The inter-annotator agreement calculated by Fleiss' Kappa score of the dataset is 81.80\%. Also, the dataset is split into the training, development, and test sets with proportion 7-1-2. Table \ref{tab:dataset} shows the overview information about the dataset.

\begin{table}[H]
\centering
\resizebox{\columnwidth}{!}{\begin{tabular}{lrrrr}
\hline
\multicolumn{1}{c}{\textbf{}} &
  \multicolumn{1}{c}{\textbf{INFORMATIVE}} &
  \multicolumn{1}{c}{\textbf{UNINFORMATIVE}}  \\ \hline
Training &            3,303          & 3,697		\\ 
Development       & 472          & 528\\ 
Test    & 944          & 1,056        \\ \hline
\end{tabular}}
\caption{Overview of the WNUT-2020 Task 2 dataset.}
\label{tab:dataset}
\end{table}

\section{Methodologies}
\label{methodologies}
In this paper, we propose an ensemble method that combines the deep learning models with the transfer learning models to identify information about COVID-19 from users' tweets. 

\subsection{Deep neural model}
\begin{figure}[H]
\centering
  \includegraphics[scale=0.28]{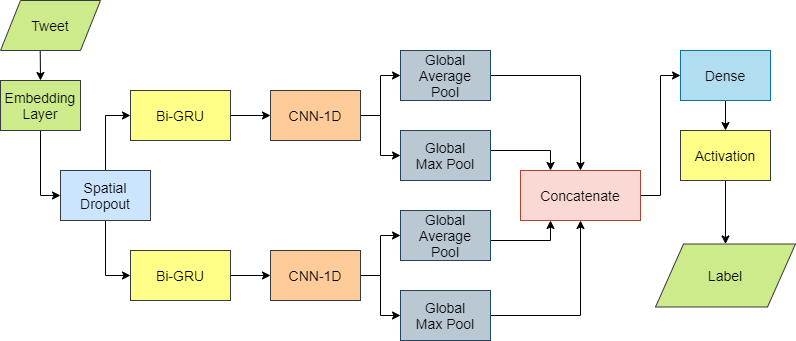}
  \caption{ Overview architecture of Bi-GRU-CNN model.}
  \label{fig:bigrucnn}
\end{figure}

We implement the Bi-GRU-CNN model, which was used for salary prediction by \citet{wang2019salary} and Job prediction by  \citet{van2020job}, with the GloVe-300d word embedding \cite{pennington2014glove}. This model consists of three main layers: the word representation layers (word embedding), the 1D Convolutional layers (CONV-1D), and the bidirectional GRU layer (Bi-GRU). This model also achieved high performances on previous study works\cite{wang2019salary, van2019hate,van2020job}. Fig \ref{fig:bigrucnn} illustrates the Bi-GRU-CNN model. 

\subsection{Transfer learning model}

Inspired by transfer learning success on many NLP tasks such as text classification \cite{do2006transfer,rizoiu2019transfer} and machine reading comprehension \cite{devlin2019,van2020new}. In this paper, we used the SOTA transfer learning models, such as BERT \cite{devlin2019}, RoBERTa \cite{liu2019roberta}, and XLNet \cite{NIPS2019_8812} with fine-tuning techniques for the problem of identifying informative tweet about COVID-19. In our experiment, we used the pre-trained language model, as described in Table \ref{tab:pre-trainedmodel}. All of these pre-trained models are constructed on English texts.

\begin{table}[H]
\begin{tabular}{p{3.8cm}p{3cm}}
\hline
\textbf{Transfer model} & \textbf{Pre-trained model} \\
\hline
BERT & \textit{bert\_en\_uncased} \\

RoBERTa & \textit{roberta-base} \\

XLNet & \textit{xlnet-large-cased} \\
\hline
\end{tabular}
\caption{List of transformer models used in our experiment.}
\label{tab:pre-trainedmodel}
\end{table}

\subsection{Ensemble method}
\begin{figure}[H]
\centering
  \includegraphics[scale=0.5]{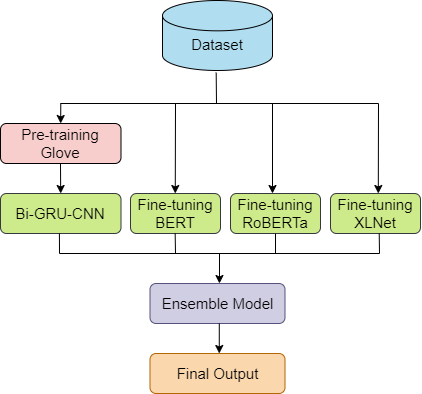}
  \caption{ Overview architecture of our ensemble approach.}
  \label{fig:ensemblemodel}
\end{figure}

As the success of the ensemble models of previous tasks \cite{van2020job,nguyennlp}, we propose a simple yet effective ensemble approach with the majority voting between the outputs of four different models, including Bi-GRU-CNN, BERT, RoBERTa, and XLNet for classifying whether a tweet contains information about COVID-19 or not. Fig \ref{fig:ensemblemodel} describes our ensemble model.

\section{Experiment}
\label{experiment}
\subsection{Experimental settings}
In this study, we experimented with datasets provided by WNUT-2020 Task 2. Training, development, and testing sets are divided as described in Section \ref{dataset}. To evaluate our models, we use four metrics include accuracy, precision, recall, and F1.

To prepare data for the model training and model evaluation phases, we perform the simple and effective pre-processing of input data as follows:

\begin{itemize}
    \item Step 1: Converting the tweet into the lowercase strings.
    \item Step 2: Removing the user names in the tweets.
    \item Step 3: Deleting all URLs in the tweets.
    \item Step 4: Representing words into vectors with pre-trained word embedding sets for deep neural network models.
\end{itemize}

According to analyzing the length of the tweets in the data, we set max\_length of the models to be 512 and epochs to be 15 for two models Bi-GRU-CNN and XLNet, and 3 for model BERT and RoBERTa. After searching for extensive hyper-parameter, we set learning\_rate equal to 1e-3 and dropout equal to 0.2 for the Bi-GRU-CNN model and learning\_rate equal to 1e-5 and dropout equal to 0.1 for three models BERT, RoBERTa, and XLNet.

\subsection{Experimental results}
Experimental results of the single model and the ensemble model on the development set are presented in Table \ref{tab:results}. Specifically, in the single models, the Bi-GRU-CNN model gives the lowest performance with 85.66\% by F1 and 86.10\% by accuracy. The single model with the highest efficiency is XLNet, which attained 89.86\% by F1 and 90.30\% by accuracy. In addition, the BERT model gives the highest Precision with 89.53\%, and the RoBERTa model achieved the highest Recall result with 90.74\%. In particular, our recommend ensemble model gives the best performance when combining the power of single models together, which accomplished 90.65\%, 91.00\%, and 92.37\% by F1, Accuracy, and Recall respectively, according to Table \ref{tab:results}. Specifically, our model improved 0.79\% by F1 and 0.70\% by Accuracy over the most extensive single model (XLNet), and 0.63\% by Recall over the RoBERTa model.

\begin{table}[H]
\centering

\resizebox{\columnwidth}{!}{\begin{tabular}{lrrrr}
\hline
\multicolumn{1}{c}{\textbf{Model}} &
  \multicolumn{1}{c}{\textbf{Accuracy}} &
  \multicolumn{1}{c}{\textbf{Precision}} &
  \multicolumn{1}{c}{\textbf{Recall}} &
  \multicolumn{1}{c}{\textbf{F1}} \\ \hline
Bi-GRU-CNN & 86.10          & 83.50          & 87.92          & 85.66          \\ 
BERT       & 89.79          & \textbf{89.53}          & 88.77          & 89.15          \\ 
RoBERTa    & 89.90          & 87.47          & \textbf{91.74}          & 89.56          \\ 
XLNet      & \textbf{90.30}          & 88.66          & 91.10          & \textbf{89.86}          \\ 
Ensemble   & \textbf{91.00} & \textbf{88.98} & \textbf{92.37} & \textbf{90.65} \\ \hline
\end{tabular}}
\caption{Model performances on the development set of the COVID-19 dataset}
\label{tab:results}
\end{table}

After the system evaluation of WNUT-2020 Task 2, Table \ref{tab:comparetop5} displays our ensemble model results on the testing set. This result is compared with the top 5 highest teams' results and the baseline model (BASELINE - FASTTEXT). Our model with F1 is 88.81\%, 2.15\% lower than the first rank team, and 13.78\% higher than the baseline model. As for the results of accuracy, we get 89.40\%, 2.10\% lower than the first rank team, and 12.10\% higher than the baseline model.

\begin{table}[H]

\resizebox{\columnwidth}{!}{\begin{tabular}{lrrrr}
\hline
\multicolumn{1}{c}{\textbf{Team Name}} &
  \multicolumn{1}{c}{\textbf{Accuracy}} &
  \multicolumn{1}{c}{\textbf{Precision}} &
  \multicolumn{1}{c}{\textbf{Recall}} &
  \multicolumn{1}{c}{\textbf{F1}} \\ \hline
NutCracker          & 91.50          & 91.35          & 90.57          & 90.96          \\ 
NLP\_North          & 91.40          & 90.29          & 91.63          & 90.96          \\ 
SupportNUTMachine   & 91.40          & 90.46          & 91.42          & 90.94          \\ 
\#GCDH              & 91.25          & 89.19          & 92.69          & 90.91          \\ 
Loner               & 91.20          & 89.18          & 92.58          & 90.85          \\
BASELINE - FASTTEXT & 77.30          & 72.88          & 77.10          & 75.03          \\ 
\textbf{BANANA}     & \textbf{89.40} & \textbf{88.53} & \textbf{89.09} & \textbf{88.81} \\ \hline
\end{tabular}}
\caption{Comparison with our model's performance with that of other teams.}
\label{tab:comparetop5}
\end{table}

\subsection{Result analysis}
\begin{figure}[H]
\centering
  \includegraphics[scale=0.3]{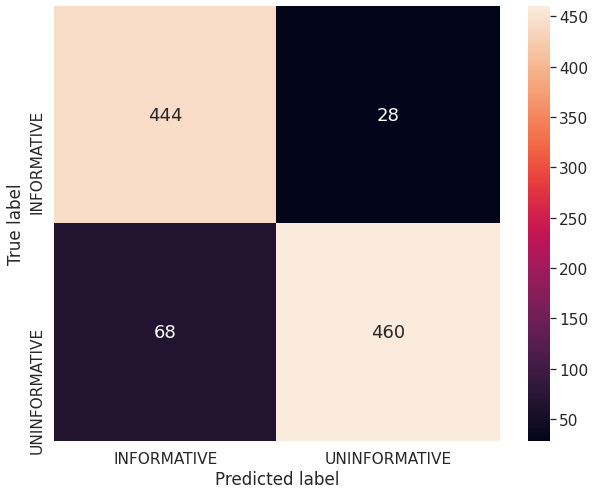}
  \caption{Confusion matrix of our ensemble model on the COVID English Tweet dataset.}
  \label{fig:conf_matrix}
\end{figure}

Fig. \ref{fig:conf_matrix} describes the confusion matrix of the ensemble methods when predicting informative tweets about COVID-19. It can be inferred from Fig. \ref{fig:conf_matrix} that the ability of prediction correct label on the INFORMATIVE label is higher than the UNINFORMATIVE label. 

In addition, Table \ref{tab:error_example} displays some error prediction examples from the dataset. Most of the wrong predictions occurred because of the appearance of special characters such as hashtag, the \textbf{HTTPURL} phrases, which stand for the URL links in the tweets. For the INFORMATION tweets, the appearance of the \textbf{HTTPURL} phrase and the hashtag \textbf{\#coronavirus} make the classification model predict the wrong label. This mistake is the same for the UNINFORMATION tweets, where the appearance of \textbf{HTTPURL} phrase and the hashtags related to the Coronavirus affected the results of the prediction model. 

\begin{table}[H]
\begin{subtable}{1\linewidth}
\label{tab:errorINFOR}
 \footnotesize
\begin{tabular}{p{0.7\textwidth}p{0.075\textwidth}|p{0.075\textwidth}|}

\hline
\multicolumn{1}{c}{\textbf{Tweet}} &
  \multicolumn{1}{c}{\textbf{PL}} &
  \multicolumn{1}{c}{\textbf{TL}} \\ \hline
NATIONAL NEWS: Coronavirus: Linda Lusardi says COVID-19 made her want to die and turned her vomit blue HTTPURL &
  \multicolumn{1}{c}{UN} &
  \multicolumn{1}{c}{IN} \\ \hline
Oh, he was sick before getting on the flight back to Australia. So Vail, Denver and possibly LA (if the layover was long enough) are suspect. I bet the LA airport is where he got it :( \#coronavirus &
  \multicolumn{1}{c}{UN} &
  \multicolumn{1}{c}{IN} \\ \hline
20/03/20 PRESS RELEASE Council Urges Consumers to be Considerate Fears about a positive coronavirus (COVID-19) case led shoppers in Fiji to begin stocking up on supplies to fill pantries. HTTPURL &
  \multicolumn{1}{c}{UN} &
  \multicolumn{1}{c}{IN} \\ \hline
\end{tabular}
\caption{Misclassified INFORMATION examples.}
\end{subtable}%

\begin{subtable}{1\linewidth}
\label{tab:errorUNINFOR}\footnotesize
\begin{tabular}{p{0.7\textwidth}p{0.075\textwidth}|p{0.075\textwidth}|}
\hline
\multicolumn{1}{c}{\textbf{Tweet}} &
  \multicolumn{1}{c}{\textbf{PL}} &
  \multicolumn{1}{c}{\textbf{TL}} \\ \hline
Vegetable market Peshawar KP.! People here are least worried about \#COVID19 .! 1 infected Asymptomatic person and he would be transmitting it to complete city inside and outside ! HTTPURL &
  \multicolumn{1}{c}{IN} &
  \multicolumn{1}{c}{UN} \\ \hline
FYI, the state's \#COVID19 stats show 395 people have been tested via state lab. But doesn't show a total for all labs. HTTPURL &
  \multicolumn{1}{c}{IN} &
  \multicolumn{1}{c}{UN} \\ \hline
New post (Mike Pence Celebrates Story of Great Great Grandmother Recovering from Coronavirus) has been published on Randy Salars News And Comment - HTTPURL HTTPURL &
  \multicolumn{1}{c}{IN} &
  \multicolumn{1}{c}{UN} \\ \hline
\end{tabular}
\caption{Misclassified UNINFORMATION examples.}
\end{subtable}%

\caption{Misclassified examples. PL: predicted label, TL: true label}
\label{tab:error_example}
\end{table}

\section{Conclusion and future work}
\label{conclusion}
This paper has addressed our work on the WNUT-2020 Task 2: Identifying COVID-19 Information on Twitter. We proposed our ensemble model combining the deep learning models and the transfer learning models for detecting information about COVID-19 from users' tweets. Our ensemble model achieved 91.00\% by accuracy and 90.65\% by F1 on the development set, and achieved 89.04\% by accuracy and 88.81\% by F1 on the public test set, which ranked \#25 in the competition.

In the future, we will improve our model's performance by exploring different features of the users' tweets and transfer learning  models with fine-tuning techniques. Finally, we hope our study can be applied in practice for detecting COVID-19 from social networks to support the COVID-19 battle all over the world.

\bibliographystyle{acl_natbib}
\bibliography{emnlp2020}

\end{document}